\documentclass{article}

\usepackage[final]{corl_2019} 

\usepackage{epsfig}
\usepackage{graphicx}
\usepackage{amsmath}
\usepackage{amssymb}
\usepackage{gensymb}

\usepackage[utf8]{inputenc} 
\usepackage[T1]{fontenc}    
\usepackage{url}            
\usepackage{booktabs}       
\usepackage{amsfonts}       
\usepackage{nicefrac}       
\usepackage{microtype}      
\usepackage{bbm}

\usepackage{amssymb}
\usepackage{subcaption}
\usepackage{changepage}
\usepackage{epstopdf}
\usepackage{xcolor}
\usepackage{colortbl}
\definecolor{LightCyan}{rgb}{0.8,1,1}
\usepackage{mathtools}
\usepackage{amsmath}
\usepackage{setspace}
\usepackage{booktabs}
\usepackage{array}
\usepackage{algorithm}
\usepackage{algorithmic}
\usepackage{multirow}

\newcommand{\colvec}[1]{\begin{bmatrix}#1\end{bmatrix}}

\def\bal#1\eal{\begin{align*}#1\end{align*}}
\newcommand{\R}{\mathbb{R}}
\newcommand{\T}{\intercal}

\newcommand{\removed}[1]{}

\newcommand{\presec}{\vspace*{0mm}}
\newcommand{\postsec}{\vspace*{0mm}}
\newcommand{\presubsec}{\vspace*{0pt}}
\newcommand{\postsubsec}{\vspace*{0pt}}
\newcommand{\presubsubsec}{\vspace*{0pt}}
\newcommand{\postsubsubsec}{\vspace*{0pt}}
\newcommand{\prepar}{\vspace*{-0pt}}
\newcommand{\preeq}{\vspace*{-0mm}}
\newcommand{\posteq}{\vspace*{-0mm}}

\title{The Best of Both Modes: Separately Leveraging RGB and Depth for Unseen Object Instance Segmentation}

\author{Christopher Xie$^1$ \hspace{4px} Yu Xiang$^2$ \hspace{4px} Arsalan Mousavian$^2$ \hspace{4px} Dieter Fox$^{2,1}$\\
$^1$University of Washington \hspace{6px} $^2$NVIDIA\\
\tt\small chrisxie@cs.washington.edu \hspace{2px} \{yux,amousavian,dieterf\}@nvidia.com
}

\begin{document}
\maketitle


\begin{abstract}
In order to function in unstructured environments, robots need the ability to recognize unseen novel objects. We take a step in this direction by tackling the problem of segmenting unseen object instances in tabletop environments. However, the type of large-scale real-world dataset required for this task typically does not exist for most robotic settings, which motivates the use of synthetic data. We propose a novel method that separately leverages synthetic RGB and synthetic depth for unseen object instance segmentation. Our method is comprised of two stages where the first stage operates only on depth to produce rough initial masks, and the second stage refines these masks with RGB. Surprisingly, our framework is able to learn from synthetic RGB-D data where the RGB is non-photorealistic. To train our method, we introduce a large-scale synthetic dataset of random objects on tabletops. We show that our method, trained on this dataset, can produce sharp and accurate masks, outperforming state-of-the-art methods on unseen object instance segmentation. We also show that our method can segment unseen objects for robot grasping. Code, models and video can be found at \url{https://rse-lab.cs.washington.edu/projects/unseen-object-instance-segmentation/}.
\end{abstract}


\section{Introduction}

\presec

For a robot to work in an unstructured environment, it must have the ability to recognize new objects that have not been seen before by the robot. Assuming every object in the environment has been modeled is infeasible and impractical. Recognizing unseen objects is a challenging perception task since the robot needs to learn the concept of ``objects'' and generalize it to unseen objects. Building such a robust object recognition module is valuable for robots interacting with objects, such as performing different manipulation tasks. A common environment in which manipulation tasks take place is on tabletops. Thus, in this paper, we approach this by focusing on the problem of unseen object instance segmentation (UOIS), where the goal is to separately segment every arbitrary (and potentially unseen) object instance, in tabletop environments. 


Training a perception module requires a large amount of data. In order to ensure the generalization capability of the module to recognize unseen objects, we need to learn from data that contains many various objects. However, in many robot environments, large-scale datasets with this property do not exist. Since collecting a large dataset with ground truth annotations is expensive and time-consuming, it is appealing to utilize synthetic data for training, such as using the ShapeNet repository which contains thousands of 3D shapes of different objects \cite{shapenet2015}. However, there exists a domain gap between synthetic data and real world data. Training directly on synthetic data only usually does not work well in the real world \cite{zhang2015towards}.

Consequently, recent efforts in robot perception have been devoted to the problem of Sim-to-Real, where the goal is to transfer capabilities learned in simulation to real world settings. For instance, some works have used domain adaptation techniques to bridge the domain gap when unlabeled real data is available \cite{tzeng2015adapting, bousmalis2018using}. Domain randomization \cite{tobin2017domain} was proposed to diversify the rendering of synthetic data for training. These methods mainly use RGB as input, and while that is desirable since there is evidence that models trained on (real world) RGB have been shown to produce sharp and accurate masks \cite{he2017mask}, this complicates the Sim-to-Real problem because state-of-the-art simulators typically cannot produce photo-realistic renderings. On the other hand, models trained with synthetic depth have been shown to generalize reasonably well (without fine-tuning) for simple settings such as bin-picking \cite{mahler2017dex, danielczuk2018segmenting}. However, in more complex settings, noisy depth sensors can limit the application of such methods. An ideal method would combine the generalization capability of training on synthetic depth and the ability to produce sharp masks by training on RGB.


In this work, we investigate how to utilize synthetic RGB-D images for UOIS in tabletop environments. We show that simply combining synthetic RGB images and synthetic depth images as inputs does not generalize well to the real world. To tackle this problem, we propose a simple two-stage framework that separately leverages the strengths of RGB and depth for UOIS. Our first stage is a Depth Seeding Network (DSN) that operates only on depth to produce rough initial segmentation masks. Training DSN with depth images allows for better generalization to the real world data. However, these initial masks from DSN may contain false alarms or inaccurate object boundaries due to depth senor noise. In these cases, utilizing the textures in RGB images can significantly help. 

Thus, our second stage is a Region Refinement Network (RRN) that takes an initial mask of an object from DSN and an RGB image as input and outputs a refined mask. Our surprising result is that, conditioned on initial masks, our RRN can be trained on non-photorealistic synthetic RGB images without any of the domain randomization or domain adaptation approaches of Sim-to-Real. We posit that mask refinement is an easier problem than directly using RGB as input to produce instance masks. We empirically show robust generalization across many different objects in cluttered real world data. In fact, as we show in our experiments, our RRN works almost as well as if it were trained on real data. Our framework, including the refinement stage, can produce sharp and accurate masks even when the depth sensor reading is noisy. We show that it outperforms state-of-the-art methods trained using any combination of RGB and depth as input, including Mask-RCNN \cite{he2017mask}.


To train our method, we introduce a synthetic dataset of tabletop objects in house environments. The dataset consists of indoor scenes of random ShapeNet \cite{shapenet2015} objects on random tabletops. We use a physics simulator \cite{coumans2019} to generate the scenes and render depth and non-photorealistic RGB. Despite this, training our proposed method on this dataset results in state-of-the-art results on the OCID dataset \cite{suchi2019easylabel} and the OSD dataset \cite{richtsfeld2012segmentation} introduced for UOIS.

This paper is organized as follows. After reviewing related work, we discuss our proposed method. We then describe our generated synthetic dataset, followed by experimental results and a conclusion.

\postsec
\section{Related Works}
\presec

\prepar
\paragraph{Object Instance Segmentation.} Object instance segmentation is the problem of segmenting every object instance in an image. Many approaches for this problem involve top-down solutions that combine segmentation with object proposals in the form of bounding boxes \cite{he2017mask, Chen_2018_CVPR}. However, when the bounding boxes contain multiple objects (e.g. heavy clutter in robot manipulation setups), the true segmentation mask is ambiguous and these methods struggle. More recently, a few methods have investigated bottom-up methods which assign pixels to object instances \cite{de2017semantic, Neven_2019_CVPR, Novotny_2018_ECCV}. Most of these algorithms provide instance masks with category-level semantic labels, which do not generalize to unseen objects in novel categories.

One approach to adapting object instance segmentation techniques to unseen objects is to employ ``class-agnostic'' training, which treats all object classes as one foreground category. One family of methods exploits motion cues with class-agnostic training in order to segment arbitrary moving objects \cite{xie2019object, dave2019towards}. Another family of methods are class-agnostic object proposal algorithms \cite{DeepMask, SharpMask, kuo2019shapemask}. However, these methods will segment everything and require some post-processing method to select the masks of interest. We also train our proposed method in a class-agnostic fashion, but instead focus our notion of unseen objects in particular environments such as tabletop settings.

\prepar
\paragraph{Sim-to-Real Perception.} Training a model on synthetic RGB and directly applying it to real data typically fails \cite{zhang2015towards}. Many methods employ some level of rendering randomization \cite{xiang2017posecnn, tremblay2019deep, li2017deepim, tobin2017domain, pinto2017asymmetric, sadeghi2017cadrl}, including lighting conditions and textures. However, they typically assume specific object instances and/or known object models. Another family of methods employ domain adaptation to bridge the gap between simulated and real images \cite{tzeng2015adapting, bousmalis2018using}. Algorithms trained on depth have been shown to generalize reasonably well for simple settings \cite{mahler2017dex, danielczuk2018segmenting}. However, noisy depth sensors can limit the application of such methods. Our proposed method is trained purely on (non-photorealistic) synthetic RGB-D data and is accurate even when depth sensors are inaccurate, and can be trained without adapting or randomizing the synthetic RGB.

\postsec
\section{Method}
\presec

\begin{figure*}[t]
\begin{center}
\includegraphics[width=\linewidth]{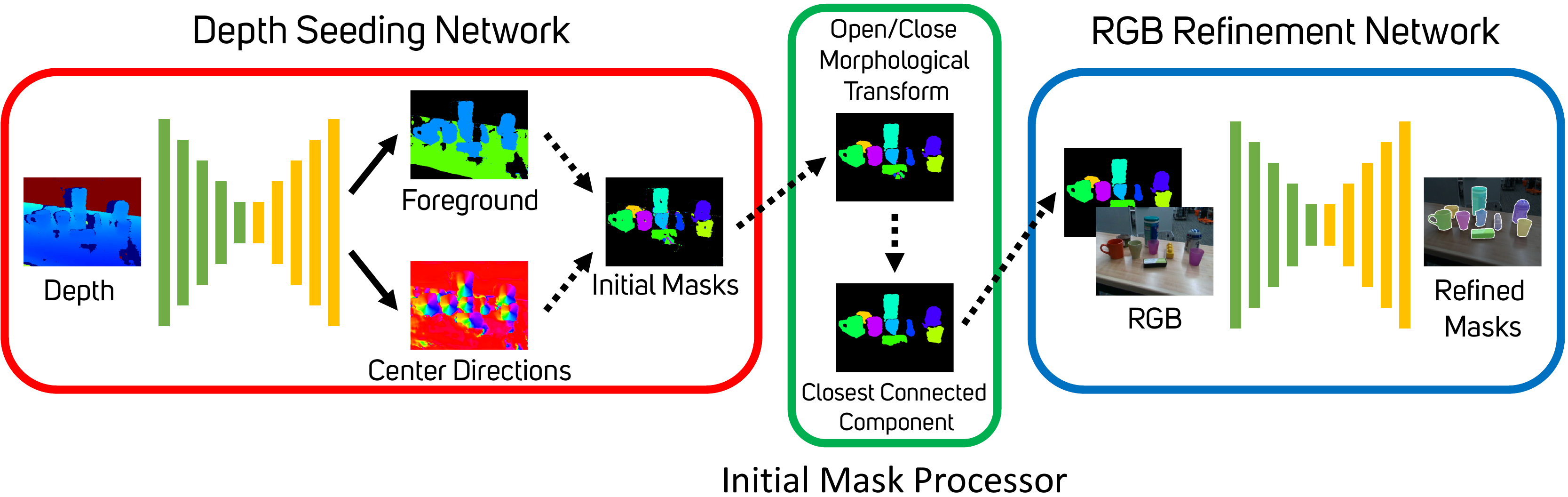}
\caption{Overall architecture. The Depth Seeding Network (\textcolor{red}{DSN}) is shown in the red box, the Initial Mask Processor (\textcolor{green}{IMP}) in the green box, and the Region Refinement Network (\textcolor{blue}{RRN}) in the blue box. The images come from a real example taken by an RGB-D camera in our lab. Despite the level of noise in the depth image (due to reflective table surface), our method is able to produce sharp and accurate instance masks. \removed{Gradients do not flow backwards through dotted lines.}}
\label{fig:architecture}
\end{center}
\vspace{-7mm}
\end{figure*}

Our framework consists of two separate networks that process Depth and RGB separately to produce instance segmentation masks. First, the Depth Seeding Network (DSN) takes an organized point cloud and outputs a semantic segmentation and 2D directions to object centers. From these, we calculate initial instance segmentation masks with a Hough voting layer. These initial masks are expected to be quite noisy, so we use an Initial Mask Processor (IMP) to robustify the masks with standard image processing techniques. Lastly, we refine the processed initial masks using our Region Refinement Network (RRN). Note that our networks, DSN and RRN, are trained separately as opposed to end-to-end. The full architecture is shown in Figure \ref{fig:architecture}.

\subsection{Depth Seeding Network}
\presubsec


\subsubsection{Network Architecture}
\presubsubsec

The DSN takes as input a 3-channel organized point cloud, $D \in \R^{H \times W \times 3}$, of XYZ coordinates. $D$ is passed through an encoder-decoder architecture to produce two outputs: a semantic segmentation mask $F \in \R^{H \times W \times C}$, where $C$ is the number of semantic classes, and 2D directions to object centers $V \in \R^{H \times W \times 2}$. We use $C=3$ for our semantic classes: background, tabletop (table plane), and tabletop objects. Each pixel of $V$ encodes a 2-dimensional unit vector pointing to the 2D center of the object. We define the center of the object to be the mean pixel location of the observable mask. Although we do not explicitly make use of the tabletop label in Section \ref{sec:experiments}, it can be used in conjunction with RANSAC \cite{fischler1981random} in order to better estimate the table and get rid of false positive masks. For the encoder-decoder architecture, we use a U-Net \cite{ronneberger2015u} architecture where each $3 \times 3$ convolutional layer is followed by a GroupNorm layer \cite{wu2018group} and ReLU. The number of output channels of this network is 64. Sitting on top of this is two parallel branches of convolutional layers that produce the foreground mask and center directions. While we use U-Net for the DSN architecture, our framework is not limited to this and can replace it with any network architecture.

In order to compute the initial segmentation masks from $F$ and $V$, we design a Hough voting layer similar to \cite{xiang2017posecnn}. First, we discretize the space of all 2D directions into $M$ equally spaced bins. For every pixel in the image, we compute the percenteage of discretized directions from all other pixels that point to it and use this as a score for how likely the pixel is an object center. We then threshold the scores to select object centers and apply non-maximum suppression. Given these object centers, each pixel in the image is assigned to the closest center it points to, which gives the initial masks as shown in the red box of Figure \ref{fig:architecture}.

\postsubsubsec
\subsubsection{Loss Function}
\presubsubsec

To train the DSN, we apply two different loss functions on the semantic segmentation $F$ and the direction prediction $V$. For the semantic segmentation loss, we use a weighted cross entropy as this has been shown to work well in detecting object boundaries in imbalanced images \cite{xie2015holistically}. The loss is 
$
\ell_s = \sum_i w_i\ \ell_{CE}\left(\hat{F}_i, F_i\right)$
where $i$ ranges over pixels, $\hat{F}_i, F_i$ are the predicted and ground truth probabilities of pixel $i$, respectively, and $\ell_{CE}$ is the cross-entropy loss. The weight $w_i$ is inversely proportional to the number of pixels with labels equal to $F_i$, normalized to sum to 1.

We apply a weighted cosine similarity loss to the direction prediction $V$. The cosine similarity is focused on the pixels belonging to the tabletop object semantic class, but we also apply it to the background/tabletop pixels to have them point in a fixed direction to avoid potential false positives. The loss is given by 
\preeq
\[
\ell_d = \frac12 \sum_{i \in O} \alpha_i \left(1 - \hat{V}_i^\T V_i\right) + \frac{\lambda_{bt}}{2} \sum_{i \in B \cup T} \beta_i \left(1 - \hat{V}_i^\T \colvec{0 \\ 1} \right)
\]
\posteq
where $\hat{V}_i, V_i$ are the predicted and ground truth unit directions of pixel $i$, respectively. $B, T, O$ are the sets of pixels belonging to background, table, and tabletop object classes, respectively. $\alpha_i$ is inversely proportional to the number of pixels with the same \textit{instance} label as pixel $i$, giving equal weight to each instance regardless of size, while $\beta_i = \frac{1}{|B \cup T|}$. We set $\lambda_{bt} = 0.1$.

The total loss is given by $\ell_s + \ell_d$. 
\postsubsubsec
\subsection{Initial Mask Processing Module}
\presubsec

Computing the initial masks from $F$ and $V$ with the Hough voting layer often results in noisy masks (see an example in Figure \ref{fig:architecture}). For example, these instance masks often exhibit salt/pepper noise and erroneous holes near the object center. As shown in Section \ref{sec:experiments}, the RRN has trouble refining the masks when they are scattered as such. To robustify the algorithm, we propose to use two simple image processing techniques to clean the masks before refinement.

For a single instance mask, we first apply an opening operation, which consists of mask erosion followed by mask dilation \cite{serra1983image}, removing the salt/pepper noise issues. Next we apply a closing operation, which is dilation followed by erosion, which closes up small holes in the mask. Finally, we select the closest connected component to the object center and discard all other components. Note that these operations are applied to each instance mask separately.

\postsubsec
\subsection{Region Refinement Network}
\presubsec

\begin{figure*}[t]
\begin{center}
\includegraphics[width=\linewidth]{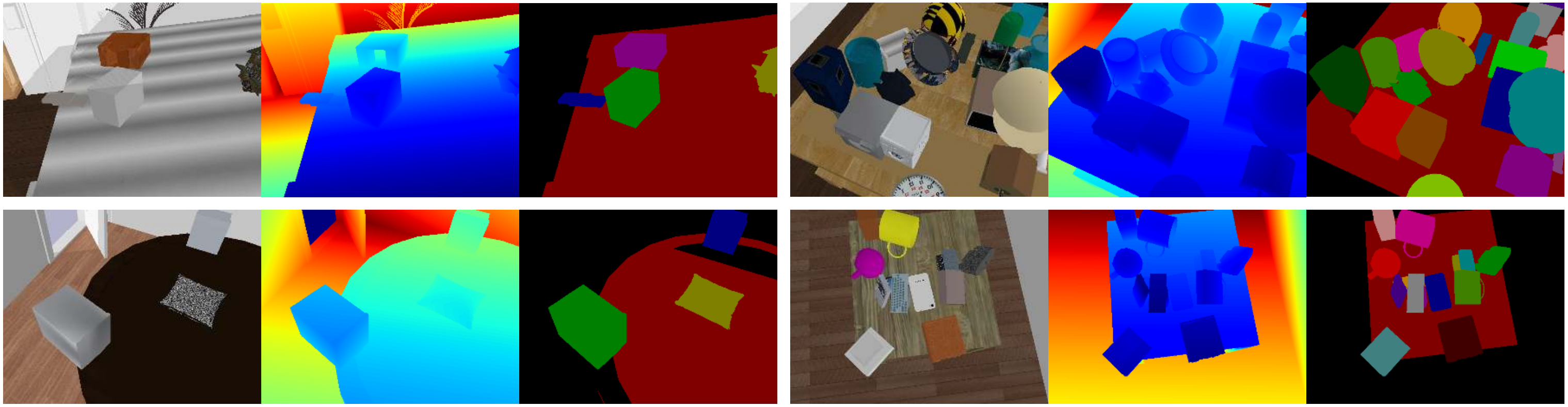}
\caption{Examples from our Tabletop Object Dataset. RGB, depth, and instance masks are shown.}
\label{fig:tod_image}
\end{center}
\vspace{-6mm}
\end{figure*}


\postsubsec
\subsubsection{Network Architecture}
\presubsubsec

This network takes as input a 4-channel image, which is a cropped RGB image concatenated with an initial instance mask. The full RGB image is cropped around the instance mask with some padding for context, concatenated with the (cropped) mask, then resized to $224 \times 224$. This gives an input image $I \in \R^{224 \times 224 \times 4}$. The output of the RRN is the refined mask probabilities $R \in \R^{224 \times 224}$, which we threshold to get the final output. We use the same U-Net architecture as in the DSN. To train the RRN, we apply the loss $\ell_s$ with two classes (foreground vs. background) instead of three.

\postsubsubsec
\subsubsection{Mask Augmentation}
\label{subsubsec:mask_augmentation}
\presubsubsec

In order to train the RRN, we need examples of perturbed masks along with ground truth masks. Since such perturbations do not exist, this problem can be seen as a data augmentation task where we augment the ground truth mask into something that resembles an initial mask (after the IMP). To this end, we detail the different augmentation techniques used to train the RRN.

\begin{itemize}
    
    \item Translation/rotation: We translate the mask by sampling a displacement vector proportionally to the mask size. \removed{from a beta distribution.} Rotation angles are sampled uniformly in $[-10\degree, 10\degree]$.
    
    \item Adding/cutting: For this augmentation, we choose a random part of the mask near the edge, and either remove it (cut) or copy it outside of the mask (add). This reflects the setting when the initial mask egregiously overflows from the object, or is only covering part of it.
    
    \item Morphological operations: We randomly choose multiple iterations of either erosion or dilation of the mask. The erosion/dilation kernel size is set to be a percentage of the mask size, where the percentage is sampled from a beta distribution. This reflects inaccurate boundaries in the initial mask, e.g. due to noisy depth sensors.
    
    \item Random ellipses: We sample the number of ellipses to add or remove in the mask from a Poisson distribution. For each ellipse, we sample both radii from a gamma distribution, and sample a rotation angle as well. This augmentation requires the RRN to learn to remove irrelevant blots outside of the object and close up small holes within it.
    
\end{itemize}


\vspace{-4mm}
\section{Tabletop Object Dataset}
\vspace{-2mm}

Many desired robot environment settings (e.g. kitchen setups, cabinets) lack large scale training sets to train deep networks. To our knowledge, there is also no large scale dataset for tabletop objects. To remedy this, we generate our own synthetic dataset which we name the Tabletop Object Dataset (TOD). This dataset is comprised of 40k synthetic scenes of cluttered objects on a tabletop in home environments. We use the SUNCG house dataset \cite{song2016ssc} for home environments and ShapeNet \cite{shapenet2015} for tables and arbitrary objects. We only use ShapeNet tables that have convex tabletops, and filter the ShapeNet object classes to roughly 25 classes of objects that could potentially be on a table. 


Each scene in the dataset is of a random table from ShapeNet inside a random room from a SUNCG house. We randomly sample anywhere between 5 and 25 objects to put on the table. The objects are either randomly placed on the table, on top of another object (stacked), or generated at a random height and orientation above the table. We use PyBullet \cite{coumans2019} to simulate physics until the objects come to rest and remove any objects that fell off the table. Next, we generate seven images (RGB, depth, and ground truth instance masks) using PyBullet's rendering capabilities. One view is of only background, another is of just the table in the room, and the rest are taken with random camera viewpoints with the tabletop objects in view. The viewpoints are sampled at a height of between .5m and 1.2m above the table and rotated randomly with an angle in $[-12\degree, 12\degree]$. The images are generated at a resolution of $640 \times 480$ with vertical field-of-view of 45 degrees. The segmentation has a tabletop (table plane, not including table legs) label and instance labels for each object.

We show some example images of our dataset in Figure \ref{fig:tod_image}. The rightmost two examples show that some of our scenes are heavily cluttered. Note that the RGB looks non-photorealistic. In particular, PyBullet is unable to load textures of some ShapeNet objects (see gray objects in leftmost two images). PyBullet was built for reinforcement learning, not computer vision, thus its rendering capabilities are insufficient for photorealistic tasks \cite{coumans2019}. Despite this, our RRN is able to learn to snap masks to object boundaries from this synthetic dataset.

\postsec
\section{Experiments}
\label{sec:experiments}
\presec

\begin{table*}[t]
\centering
\begin{tabular}{|c|ccc|ccc|}
\hline
\multirow{2}{*}{Method} & \multicolumn{3}{c|}{Overlap} & \multicolumn{3}{c|}{Boundary} \\
 & \textcolor{orange}{P} & \textcolor{cyan}{R} & \textcolor{purple}{F} & \textcolor{orange}{P} & \textcolor{cyan}{R} & \textcolor{purple}{F}\\
\hline
GCUT \cite{felzenszwalb2004efficient} & 21.5 & 51.5 & {25.7} & 10.2 & 46.8 & {15.7} \\
SCUT \cite{pham2018scenecut} & 45.7 & 72.5 & {43.7} & 43.1 & 65.1 & {42.6} \\
LCCP \cite{christoph2014object} & 58.4 & \textcolor{red}{89.1} & {63.8} & 53.6 & \textcolor{red}{82.6} & {60.2} \\
V4R \cite{potapova2014incremental} & 65.3 & 81.4 & {69.5} & 62.5 & 81.4 & {66.6} \\
Ours & \textcolor{red}{88.8} & 81.7 & {\textcolor{red}{84.1}} & \textcolor{red}{83.0} & 67.2 & {\textcolor{red}{73.3}} \\
\hline
\end{tabular}
\caption{Comparison with baselines on the ARID20 and YCB10 subsets of OCID \cite{suchi2019easylabel}.}
\label{table:baseline_comparison}
\vspace{-5mm}
\end{table*}

\subsection{Implementation Details}
\presubsec

In order to seek a fair comparison, all models trained in this section are trained for 100k iterations of SGD using a fixed learning rate of 1e-2 and batch size of 8. All images have a resolution $H = 480, W = 640$. During DSN training, we augment depth with multiplicative noise sampled from a gamma distribution, and additive Gaussian Process noise in 3D, similar to \cite{mahler2017dex}. We augment inputs to the RRN as described in Section \ref{subsubsec:mask_augmentation}. All experiments run on a NVIDIA RTX 2080ti.

\subsection{Datasets}
\presubsec

We use TOD to train all of our models. We evaluate quantitatively and qualitatively on two real-world datasets: OCID \cite{suchi2019easylabel} and OSD \cite{richtsfeld2012segmentation}, which have 2346 images of semi-automatically constructed labels and 111 manually labeled images, respectively. OSD is a small dataset that was manually annotated, so the annotation quality is high. However, the OCID dataset, which is much larger, uses a semi-automatic process of annotating the labels. It exploits the temporal order of building up a scene by calculating difference in depth between incremental images, where only one additional object is placed. However, this process is subject to the noise of the depth sensor, so while the majority of the instance label is accurate, the boundaries of the objects are fuzzy, leading to noisy instance label boundaries. Additionally, OCID contains images with objects on a tabletop, and images with objects on a floor. Despite our method being trained in synthetic tabletop settings, it generalizes to floor settings as well.


\subsection{Metrics}
\presubsec

\begin{table*}[t]
\resizebox{\linewidth}{!}{\begin{tabular}{|c|c|ccc|ccc|ccc|ccc|}
\hline
\multirow{3}{*}{Method} & \multirow{3}{*}{Input} & \multicolumn{6}{c|}{OCID \cite{suchi2019easylabel}} & \multicolumn{6}{c|}{OSD \cite{richtsfeld2012segmentation}} \\ \cline{3-14}
 &  & \multicolumn{3}{c|}{Overlap} & \multicolumn{3}{c|}{Boundary} & \multicolumn{3}{c|}{Overlap} & \multicolumn{3}{c|}{Boundary} \\
 &  & \textcolor{orange}{P} & \textcolor{cyan}{R} & \textcolor{purple}{F} & \textcolor{orange}{P} & \textcolor{cyan}{R} & \textcolor{purple}{F} & \textcolor{orange}{P} & \textcolor{cyan}{R} & \textcolor{purple}{F} & \textcolor{orange}{P} & \textcolor{cyan}{R} & \textcolor{purple}{F} \\ \hline
Mask RCNN & RGB & 66.0 & 34.0 & 36.6 & 58.2 & 25.8 & 29.0 & 63.7 & 43.1 & 46.0 & 46.7 & 26.3 & 29.5 \\
Mask RCNN & Depth & 82.7 & \textcolor{red}{78.9} & 79.9 & 79.4 & \textcolor{red}{67.7} & \textcolor{red}{71.9} & 73.8 & 72.9 & 72.2 & 49.6 & 40.3 & 43.1  \\
Mask RCNN & RGB-D & 79.2 & 78.6 & 78.0 & 73.6 & 67.2 & 69.2 & 74.0 & 74.6 & 74.1 & 57.3 & 52.1 & 53.8 \\
Ours & DSN: RGB-D & 84.2 & 57.6 & 62.2 & 72.9 & 44.5 & 49.9 & 72.2 & 63.7 & 66.1 & 58.5 & 43.4 & 48.4 \\
Ours & DSN: Depth & \textcolor{red}{88.3} & \textcolor{red}{78.9} & \textcolor{red}{81.7} & \textcolor{red}{82.0} & 65.9 & 71.4 & \textcolor{red}{80.7} & \textcolor{red}{80.5} & \textcolor{red}{79.9} & \textcolor{red}{66.0} & \textcolor{red}{67.1} & \textcolor{red}{65.6} \\
\hline

\end{tabular}}
\caption{Evaluation of our method against state-of-the-art instance segmentation algorithm Mask RCNN trained on different input modes.}
\label{table:input_comparison_sota}
\vspace{-1mm}
\end{table*}

We use the precision/recall/F-measure (P/R/F) metrics as defined in \cite{dave2019towards}. This metric promotes methods that segment the desired objects, however they penalize methods that provide false positives. Specifically, the precision, recall, and F-measure are computed between all pairs of predicted objects and ground truth objects. The Hungarian method with pairwise F-measure is used to compute a matching between predicted objects and ground truth. Given this matching, the final P/R/F is computed by
\preeq
\[
P = \frac{\sum_i \left|c_i \cap g(c_i) \right|}{\sum_i \left|c_i\right|},\ \ \ R = \frac{\sum_i \left|c_i \cap g(c_i) \right|}{\sum_j \left|g_j\right|},\ \ \ F = \frac{2PR}{P+R}
\]
\posteq
where $c_i$ denotes the set of pixels belonging to predicted object $i$, $g(c_i)$ is the set of pixels of the matched ground truth object of $c_i$, and $g_j$ is the set of pixels for ground truth object $j$. We denote this as Overlap P/R/F. See \cite{dave2019towards} for more details. 

While the above metric is quite informative, it does not take object boundaries into account. To remedy this, we introduce a Boundary P/R/F measure to complement the Overlap P/R/F. To compute Boundary P/R/F, we use the same Hungarian matching used to compute Overlap P/R/F. Given these matchings, the Boundary P/R/F is computed by
\preeq
\[
P = \frac{\sum_i \left|c_i \cap D\left[g(c_i)\right] \right|}{\sum_i \left|c_i\right|},\ \ \ R = \frac{\sum_i \left|D\left[c_i\right] \cap g(c_i) \right|}{\sum_j \left|g_j\right|},\ \ \ F = \frac{2PR}{P+R}
\]
\posteq
where we overload notation and denote $c_i, g_j$ to be the set of pixels belonging to the boundaries of predicted object $i$ and ground truth object $j$, respectively. $D[\cdot]$ denotes the dilation operation, which allows for some slack in the prediction. Roughly, these metrics are a combination of the $\mathcal{F}$-measure in \cite{Perazzi2016} along with the Overlap P/R/F as defined in \cite{dave2019towards}.

We report all P/R/F measures in the range $[0,100]$ (P/R/F $\times 100$).

\subsection{Quantitative Results}
\presubsec

\begin{table*}[t]
\centering
\resizebox{\linewidth}{!}{\begin{tabular}{|c|ccc|ccc|ccc|ccc|}
\hline
\multirow{3}{*}{\parbox{1.75cm}{\centering RRN training data}} & \multicolumn{6}{c|}{OCID \cite{suchi2019easylabel}} & \multicolumn{6}{c|}{OSD \cite{richtsfeld2012segmentation}} \\ 
\cline{2-13} & \multicolumn{3}{c|}{Overlap} & \multicolumn{3}{c|}{Boundary} & \multicolumn{3}{c|}{Overlap} & \multicolumn{3}{c|}{Boundary} \\
& \textcolor{orange}{P} & \textcolor{cyan}{R} & \textcolor{purple}{F} & \textcolor{orange}{P} & \textcolor{cyan}{R} & \textcolor{purple}{F} & \textcolor{orange}{P} & \textcolor{cyan}{R} & \textcolor{purple}{F} & \textcolor{orange}{P} & \textcolor{cyan}{R} & \textcolor{purple}{F} \\ \hline
TOD & 88.3 & 78.9 & 81.7 & 82.0 & 65.9 & 71.4 & 80.7 & 80.5 & 79.9 & 66.0 & 67.1 & 65.6 \\
OID \cite{OpenImages} & 87.9 & 79.6 & 81.7 & 84.0 & 69.1 & 74.1 & 81.2 & 83.3 & 81.7 & 69.8 & 73.7 & 70.8 \\
\hline

\end{tabular}}
\caption{Comparison of RRN when training on TOD and real images from Google OID \cite{OpenImages}.}
\label{table:RRN_real_data}
\vspace{-1mm}
\end{table*}

\begin{table*}[t]
\centering

\begin{minipage}[t]{.47\linewidth}
\resizebox{\linewidth}{!}{\begin{tabular}{|cccc|ccc|}
\hline
\multirow{2}{*}{DSN} & \multicolumn{2}{|c|}{IMP}  & \multirow{2}{*}{RRN} & \multicolumn{3}{c|}{Boundary} \\
& \multicolumn{0}{|c}{O/C} & \multicolumn{0}{c|}{CCC} & & \textcolor{orange}{P} & \textcolor{cyan}{R} & \textcolor{purple}{F} \\ \hline

\checkmark &  &  &  & 35.0 & 58.5 & 43.4 \\
\checkmark &  &  & \checkmark & 36.0 & 48.1 & 39.6 \\
\checkmark & \checkmark &  &  & 49.2 & 55.3 & 51.7 \\
\checkmark & \checkmark &  & \checkmark & 59.0 & 64.1 & 60.7 \\
\checkmark & \checkmark & \checkmark &  & 53.8 & 54.7 & 53.6\\
\checkmark & \checkmark & \checkmark & \checkmark & 66.0 & 67.1 & 65.6 \\

\hline
\end{tabular}}
\end{minipage}
\begin{minipage}[t]{.51\linewidth}
\resizebox{\linewidth}{!}{\begin{tabular}{|c|c|c|ccc|}
\hline
\multirow{2}{*}{Method} & \multirow{2}{*}{Input} & \multirow{2}{*}{RRN} & \multicolumn{3}{c|}{Boundary} \\
 &  &  & \textcolor{orange}{P} & \textcolor{cyan}{R} & \textcolor{purple}{F} \\ \hline
Mask RCNN & RGB & & 46.7 & 26.3 & 29.5 \\
Mask RCNN & RGB & \checkmark & 64.1 & 44.3 & 46.8 \\
Mask RCNN & Depth & & 49.6 & 40.3 & 43.1 \\
Mask RCNN & Depth & \checkmark & 69.0 & 55.2 & 59.8 \\
Mask RCNN & RGB-D & & 57.3 & 52.1 & 53.8 \\
Mask RCNN & RGB-D & \checkmark & 63.4 & 57.0 & 59.2\\
\hline
\end{tabular}}
\end{minipage}

\caption{(left) Ablation experiments on OSD \cite{richtsfeld2012segmentation}. O/C denotes the Open/Close morphological transform, while CCC denotes Closest Connected Component of the IMP module. (right) Refining Mask RCNN results with our RRN on OSD.}
\label{table:ablation}
\vspace{-6mm}
\end{table*}

\textbf{Comparison to baselines.} 
We compare to baselines shown in \cite{suchi2019easylabel}, which include GCUT \cite{felzenszwalb2004efficient}, SCUT \cite{pham2018scenecut}, LCCP \cite{christoph2014object}, and V4R \cite{potapova2014incremental}. In \cite{suchi2019easylabel}, these methods were only evaluated on the ARID20 and YCB10 subsets of OCID, so we compare our results this subset as well. These baselines are designed to provide over-segmentations (i.e., they segment the whole scene instead of just the objects of interest). To allow a more fair comparison, we perform the following with baseline methods results: set all predicted masks smaller than 500 pixels to background, and set the largest mask to table label (which is not considered in our metrics). Results are shown in Table \ref{table:baseline_comparison}. Because the baseline methods aim to over-segment the scene, the precision is in general low while the recall is high. LCCP is designed to segment convex objects (most objects in OCID are convex), but its predicted boundaries are noisy due to operating on depth only. Both SCUT and V4R utilize models trained on real data as part of their pipelines. V4R was trained on OSD \cite{richtsfeld2012segmentation} which has an extremely similar data distribution to OCID, giving V4R a substantial advantage. Our method, despite never having seen any real data, significantly outperforms these baselines on F-measure.

\begin{table}[h]
\centering
\begin{tabular}{|ccc|ccc|}
\hline
\multicolumn{3}{|c|}{OCID \cite{suchi2019easylabel}} & \multicolumn{3}{c|}{OSD \cite{richtsfeld2012segmentation}} \\
\textcolor{orange}{P} & \textcolor{cyan}{R} & \textcolor{purple}{F} & \textcolor{orange}{P} & \textcolor{cyan}{R} & \textcolor{purple}{F} \\ \hline
85.2 & 70.8 & 75.7 & 53.4 & 53.3 & 52.8 \\
\hline
\end{tabular}
\caption{Boundary metrics of our method without RRN refinements (DSN and IMP only).}
\label{table:OCID_issue}
\end{table}

\textbf{Effect of input mode.}
Next, we evaluate how different input modes affect the results by training Mask RCNN \cite{he2017mask}, a state-of-the-art instance segmentation algorithm, on different combinations of RGB and depth from TOD. In Table \ref{table:input_comparison_sota}, we compare Mask RCNN trained on RGB, depth, and RGB-D and compare it to our proposed model on the full OCID and OSD datasets. It is clear to see that training Mask RCNN on synthetic RGB only does not generalize at all. Concatenating depth to RGB as input boosts performance significantly. However, our method (line 5, Table \ref{table:input_comparison_sota}) exploits RGB and depth separately, leading to better results than the state-of-the-art Mask RCNN on OSD while being trained on the exact same synthetic dataset. Furthermore, we show that when our DSN is trained on RGB concatenated with depth (line 4, Table \ref{table:input_comparison_sota}), we see a drop in performance, suggesting that training directly on (non-photorealistic) RGB is not the best way of utilizing the synthetic data. 

Note that the performance of our method is similar to Mask RCNN (trained on depth and RGB-D) on OCID in terms of boundary F-measure (see Table~\ref{table:input_comparison_sota}). This result is misleading: it turns out that using the RRN to refine the initial masks result in a loss in quantitative performance on OCID, while the qualitative results are better. This is due to the semi-automatic labeling procedure in OCID that leads to ground truth segmentation labels aligning with noise from the depth camera~\cite{suchi2019easylabel}. Table~\ref{table:OCID_issue} shows our performance without applying RRN (DSN and IMP only) on boundary F-measure. This version of our method, along with Mask RCNN trained on depth, utilizes only depth and predicts segmentation boundaries that are aligned with the sensor noise. In this setting, we outperform Mask RCNN. On the other hand, OSD has accurate manually annotated labels and this issue does not hold on this dataset.

\textbf{Degradation of training on non-photorealistic simulated RGB.}
To quantify how much non-photorealistic RGB degrades performance, we train an RRN on real data. This approximately serves as an upper bound on how well the synthetically-trained RRN can perform. We use the instance masks from the Google Open Images dataset (OID) \cite{OpenImages, OpenImagesSegmentation} and filter them to relevant object classes (that might potentially be on a tabletop), resulting in approximately 220k instance masks on real RGB images. We compare our synthetically-trained RRN to an RRN trained on OID in Table \ref{table:RRN_real_data}. Both models share the same DSN and IMP. The Overlap measures are roughly the same, while the RRN trained on OID has slightly better performance on the Boundary measures. This suggests that while there is still a gap, our method is surprisingly not too far off, considering that we train the RRN with non-photorealistic synthetic RGB. We conclude that mask refinement with RGB is an easier task to transfer from Sim-to-Real than directly segmenting from RGB.

\textbf{Ablation studies.}
We report ablation studies on OSD to evaluate each component of our proposed method in Table \ref{table:ablation} (left). We omit the Overlap P/R/F results since they follow similar trends to Boundary P/R/F. Running the RRN on the raw masks output by DSN without the IMP module actually hurts performance as the RRN cannot refine such noisy masks. Adding the open/close morphological transform and/or the closest connect component results in much stronger results, showing that the IMP is key in robustifying our proposed method. In these settings, applying the RRN significantly boosts Boundary P/R/F showing that it effectively sharpens the masks. In fact, Table \ref{table:ablation} (right) shows that applying the RRN to the Mask RCNN results effectively boosts the Boundary P/R/F for all input modes, showing the efficacy of the RRN despite being directly trained on non-photorealistic RGB. Note that even with this refinement, the Mask RCNN results are outperformed by our method.



\postsubsec

\subsection{Qualitative Results}

\begin{figure*}[t]
\begin{center}
\includegraphics[width=\linewidth]{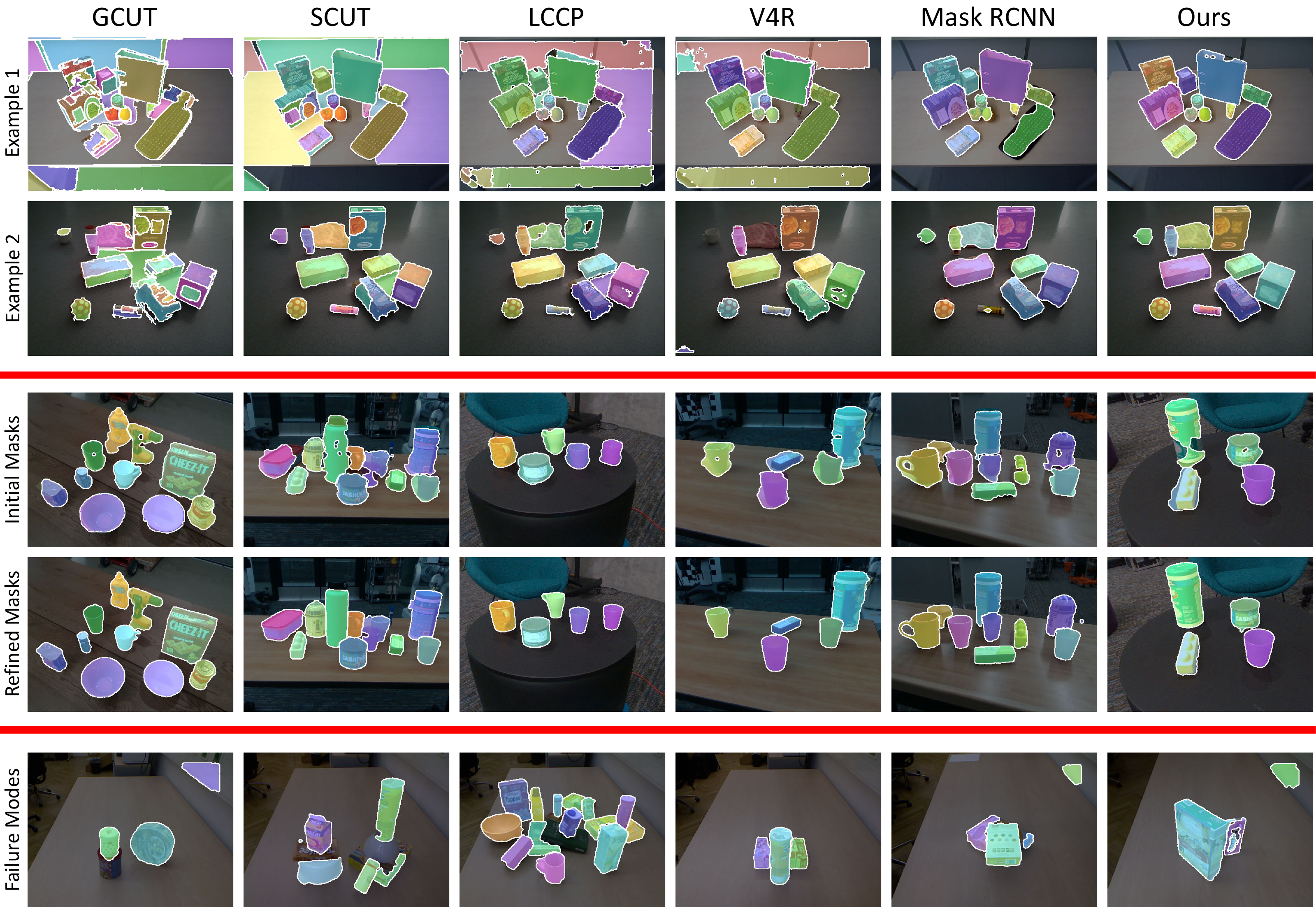}
\caption{Qualitative results. (Top) Comparison on OCID \cite{suchi2019easylabel}. (Middle) Mask refinements on images taken from our lab. (Bottom) Failure modes on OSD \cite{richtsfeld2012segmentation}.}
\label{fig:qualitative}
\end{center}
\vspace{-7mm}
\end{figure*}

We show qualitative results on OCID of baseline methods, Mask RCNN (trained on RGB-D), and our proposed method in Figure \ref{fig:qualitative} (top). It is clear that the baseline methods suffer from over-segmentation issues; they segment the table and background into multiple pieces. For the methods that utilize RGB as an input (GCUT and SCUT), the objects are often over-segmented as well. Methods that operate on depth alone (LCCP and V4R) result in noisy object boundaries due to noise in the depth sensors. The main failure mode for Mask RCNN is that it tends to undersegment objects; a close inspection of Figure \ref{fig:qualitative} shows that Mask RCNN erroneously segments multiple objects as one. This is the typical failure mode of top-down instance segmentation algorithms in clutter. On the other hand, our bottom-up method utilizes depth and rgb separately to provide sharp and accurate masks.

In Figure \ref{fig:qualitative} (middle), we qualitatively show the effect of the RRN. The first row shows the initial masks after the IMP module and the second row shows the refined masks. These images were taken around our lab with an Intel RealSense D435 RGB-D camera to demonstrate the robustness of our method to camera viewpoint variations and distracting backgrounds (OCID/OSD have relatively simple backgrounds). Due to noise in the depth sensor, it is impossible to get sharp and accurate predictions from depth alone without using RGB.
Our RRN can provide sharp masks even when the boundaries of objects are occluding other objects (images 2 and 5).

\begin{figure}
\begin{center}
\includegraphics[width=\linewidth]{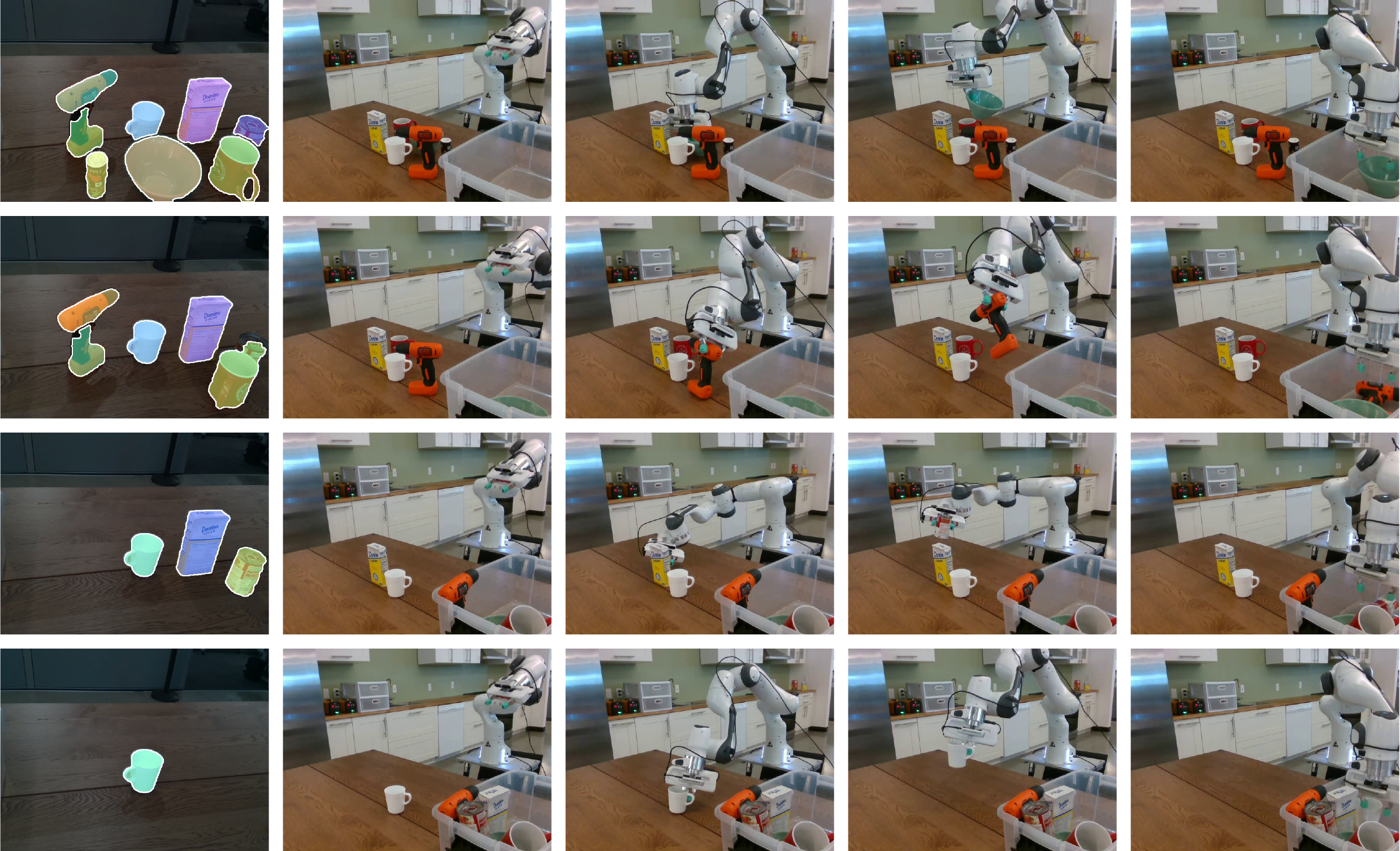}
\caption{Visualization of clearing table using our instance segmentation and 6-DOF GraspNet~\cite{mousavian2019grasp}}
\label{fig:robot}
\end{center}
\end{figure}

We show some failure modes of our method in Figure \ref{fig:qualitative} (bottom on the OSD dataset. In images 1, 5, and 6, false positives contributed by the DSN cannot be undone by the RRN. In images 2 and 3, we see examples of missed objects due to the center of the mask being occluded by an object, which is a limitation of the 2D center voting procedure. Lastly, when an object is split into two (images 3, 4, and 5), our method predicts two separate objects. See the supplemental video for more results.

\subsection{Application in Grasping Unknown Objects}
We show the application of instance segmentation in grasping unknown objects in cluttered environment using a Franka robot with panda gripper and wrist-mounted RGB-D camera. The task is to collect objects from a table and put them in a bin. Objects are segmented using our method and the point cloud of the closest object to the camera is fed to 6-DOF GraspNet \cite{mousavian2019grasp} to generate diverse grasps from the object point cloud. Other objects are represented as obstacles by sampling fixed number of points using farthest point sampling from their corresponding point cloud. The grasp that has the maximum score and has a feasible path is chosen to execute. Fig.~\ref{fig:robot} shows the instance segmentation at different stages of the task and also the execution of grasps with robot. Our method segments the object correctly most of the time but fails in two scenarios. The first one is the over-segmentation of the drill in the scene. Our method considers the top of the drill as one object and the handle as an obstacle. This is because there are missing depth value between the two parts and since the handle is in different color RRN fails to merge them back together. The second failure case is when the red mug and tomato soup can get very close to each other. In this scenario the DSN is able to distinguish these objects from each other but RRN merges them together since both objects have similar color. We conducted the experiments to collect the objects from the table 3 times and our method was able to successfully collect all the objects in each trial with 1-2 extra grasping attempts in each trial. The failures stem from either imperfections in segmentation or inaccurate generated grasps. Video of the robot experiments are included in the supplementary materials. Note that neither our instance segmentation method nor 6-DOF GraspNet are trained on real data.

\vspace{-3mm}
\section{Conclusion}
\vspace{-3mm}

We proposed a framework that separately leverages RGB and depth to provide sharp and accurate masks for unseen object instance segmentation. Our two-stage framework produces rough initial masks using only depth, then refines those masks with RGB. Surprisingly, our RRN can be trained on non-photorealistic RGB and generalize quite well to real world images. We demonstrated the efficacy of our approach on multiple datasets for UOIS in tabletop environments.

\bibliography{references}  

\end{document}